\documentclass[10pt,twocolumn,letterpaper]{article}

\usepackage[pagenumbers]{cvpr} 

\usepackage{graphicx}
\usepackage{amsmath}
\usepackage{amssymb}
\usepackage{booktabs}
\usepackage{amsfonts}
\usepackage{bbding} 
\usepackage{color}
\usepackage{multirow}

\usepackage[pagebackref,breaklinks,colorlinks]{hyperref}

\usepackage[capitalize]{cleveref}
\crefname{section}{Sec.}{Secs.}
\Crefname{section}{Section}{Sections}
\Crefname{table}{Table}{Tables}
\crefname{table}{Tab.}{Tabs.}

\def\cvprPaperID{795} 
\def\confName{CVPR}
\def\confYear{2023}

\begin{document}

\title{Are We Ready for Vision-Centric Driving Streaming Perception?
 \\The ASAP Benchmark}
\author{Xiaofeng Wang\textsuperscript{\rm 1,2}~~Zheng Zhu\textsuperscript{\rm 3}~~Yunpeng Zhang\textsuperscript{\rm 3}~~Guan Huang\textsuperscript{\rm 3}\\Yun Ye\textsuperscript{\rm 3}~~Wenbo Xu\textsuperscript{\rm 3}~~Ziwei Chen\textsuperscript{\rm 4}~~Xingang Wang\textsuperscript{\rm 1}\\
\textsuperscript{\rm 1}CASIA
~ ~ \textsuperscript{\rm 2}UCAS
~ ~ \textsuperscript{\rm 3}PhiGent Robotics
~ ~ \textsuperscript{\rm 4}SEU \\
\tt\small \{wangxiaofeng2020,xingang.wang\}@ia.ac.cn
zhengzhu@ieee.org\\
\tt\small \{yunpeng.zhang,guan.huang,yun.ye,wenbo.xu\}@phigent.ai richard\_chen@seu.edu.cn}




\maketitle

\begin{abstract}
\vspace{-0.3cm}

In recent years, vision-centric perception has flourished in various autonomous driving tasks, including 3D detection, semantic map construction, motion forecasting, and depth estimation.
Nevertheless, the latency of vision-centric approaches is too high for practical deployment (\eg, most camera-based 3D detectors have a runtime greater than 300ms).
To bridge the gap between ideal researches and real-world applications, it is necessary to quantify the trade-off between performance and efficiency.
Traditionally, autonomous-driving perception benchmarks perform the \textbf{offline} evaluation, neglecting the inference time delay. To mitigate the problem, we propose the \textbf{A}utonomous-driving \textbf{S}tre\textbf{A}ming \textbf{P}erception (ASAP) benchmark, which is the first benchmark to evaluate the \textbf{online} performance of vision-centric perception in autonomous driving.
On the basis of the 2Hz annotated nuScenes dataset, we first propose an annotation-extending pipeline to generate high-frame-rate labels for the 12Hz raw images. 
Referring to the practical deployment, the \textbf{S}treaming \textbf{P}erception \textbf{U}nder const\textbf{R}ained-computation (SPUR) evaluation protocol is further constructed, where the 12Hz inputs are utilized for streaming evaluation under the constraints of different computational resources. In the ASAP benchmark, comprehensive experiment results reveal that the model rank alters under different constraints, suggesting that the model latency and computation budget should be considered as design choices to optimize the practical deployment.
To facilitate further research, we establish baselines for camera-based streaming 3D detection, which consistently enhance the streaming performance across various hardware. ASAP project page: \url{https://github.com/JeffWang987/ASAP}.
\end{abstract}

\vspace{-0.7cm}
\section{Introduction}
\label{sec:intro}

\begin{figure}[ht]
\centering
\resizebox{1\linewidth}{!}{
\includegraphics{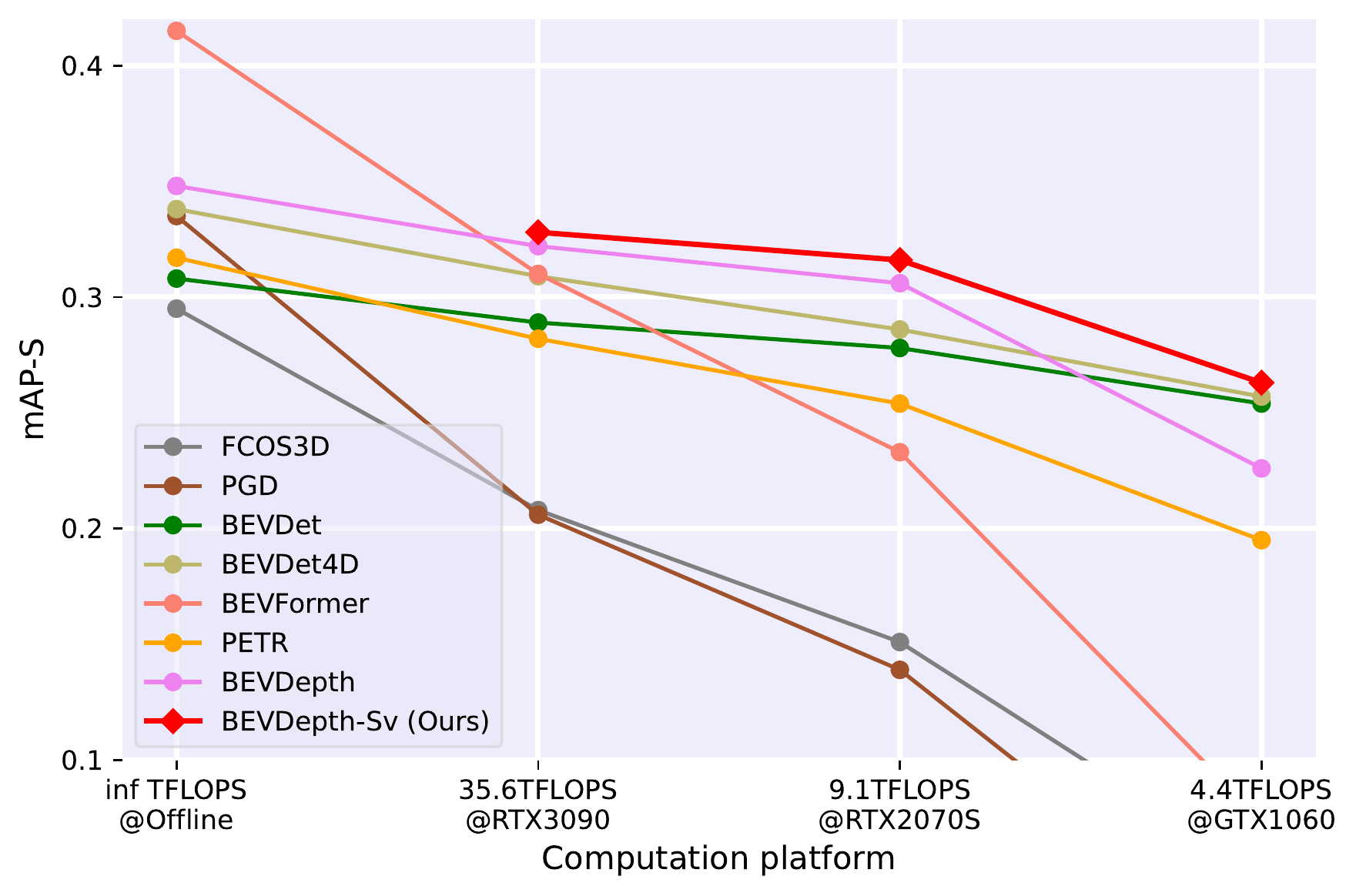}}
\caption{Comparison of streaming performances on the ASAP benchmark, where the model rank changes under different computational resources. Note that our baseline BEVDepth-Sv (built upon\cite{li2022bevdepth}) consistently improves the streaming performance on different platforms.}
\label{fig:key1}
\vspace{-0.2cm}
\end{figure}

Vision-centric perception in autonomous driving has drawn extensive attention recently, as it can obtain richer semantic information from images with a desirable budget, compared to LiDAR-based perception. 
Notably, the past years have witnessed the blooming of vision-centric perception in various autonomous driving tasks (\eg, 3D detection \cite{li2022bevformer,huang2021bevdet,huangjj2021BEVDet4D,liu2022petr,liu2022petrv2,li2022bevdepth,li2022bevstereo,en2022m2bev,Jin2022time,Zeng2022sts}, semantic map construction \cite{li2022hdmapnet,peng2022bevsegformer,WeixiangYang2021ProjectingYV,zou2022hft,pon,vpn}, motion forecasting \cite{AnthonyHu2021FIERYFI,akan2022stretchbev}, and depth estimation \cite{wang2022crafting,monov2,manydepth,wei2022surround,packnet,FelixWimbauer2020MonoRecSD}).                   

\begin{table}[ht]
  \centering
  \caption{Comparison between autonomous-driving perception dataset, where \textit{L\&C} represents LiDAR and camera, \#\textit{sensors} denotes number of sensors, \textit{Ann. frequency} is the annotation frequency, and \textit{Model speed} denotes the typical inference speed of the model on RTX3090. For 2D detectors \cite{yolox2021,yolov3,yolov4,yolov5,EfficientDet}, they achieve $\sim$45mAP@30FPS on COCO \cite{MSCOCO}. For LiDAR-based 3D detectors \cite{yin2021center,YanYan2018SECONDSE,AlexHLang2018PointPillarsFE}, they achieve $\sim$70mAP@20FPS on Waymo \cite{PeiSun2020ScalabilityIP}. For camera-based 3D detectors \cite{li2022bevdepth,huangjj2021BEVDet4D,li2022bevformer,zhang2022beverse}, they achieve $\sim$40mAP@3FPS on nuScenes \cite{HolgerCaesar2019nuScenesAM}, which is 6$\times$$\sim$10$\times$ slower than the above two tasks.}
  \resizebox{1.0\linewidth}{!}{
    \begin{tabular}{lc|ccccc}
    \hline
    \multirow{2}{*}{Dataset} & \multirow{2}*{Stream.} & \multirow{2}*{Modality}  &  \multirow{2}*{\#sensors} & \multirow{2}*{Task} & Ann. & Model \\ 
     & & & & &frequency &speed \\
    \hline
    KITTI \cite{Geiger2012CVPR}&\XSolidBrush &   L\&C   &  -   & Multi-task & - & -    \\
    Argoverse \cite{Ben2021argo} &\XSolidBrush &   L\&C   &  -   & Multi-task   & -& -     \\
    Waymo \cite{PeiSun2020ScalabilityIP} &\XSolidBrush & L\&C&   -  & Multi-task   & - & -     \\
    nuScenes \cite{HolgerCaesar2019nuScenesAM}  &\XSolidBrush& L\&C &    -  & Multi-task   & -  & -     \\
    \hline
    Argoverse-HD \cite{meng2020sap}  &\Checkmark &   C   &  1   & 2D det. & 30Hz   & $\sim$30FPS    \\
    Waymo \cite{WeiHan2020StreamingOD}  &\Checkmark &  L&  1  & L-3D det. & 10Hz  & $\sim$20FPS     \\
    nuScenes-H   &\Checkmark& C &    6  & C-3D det. & 12Hz   & $\sim$3FPS      \\
    \hline
    \end{tabular}%
    }
  \label{tab:stream_cmp}%
\end{table}%

Despite the growing research interest in vision-centric approaches, the high latency of these methods 
still prevents the practical deployment.
Specifically, in the fundamental task of autonomous-driving perception (\eg, 3D detection), the inference time of most camera-based 3D detectors \cite{li2022bevformer,huangjj2021BEVDet4D,liu2022petrv2,li2022bevdepth, li2022bevstereo,en2022m2bev,zhang2022beverse} is longer than 300ms (on the powerful NVIDIA RTX3090), which is $\sim$6$\times$ longer (see Tab.~\ref{tab:stream_cmp}) than the LiDAR-based counterparts \cite{yin2021center,YanYan2018SECONDSE,AlexHLang2018PointPillarsFE}.
To enable practical vision-centric perception in autonomous driving, a quantitative metric is 
in an urgent need
to balance the accuracy and latency. However, previous autonomous-driving benchmarks \cite{HolgerCaesar2019nuScenesAM,PeiSun2020ScalabilityIP,Geiger2012CVPR,MariusCordts2016TheCD,bdd100k,XinyuHuang2020TheAO,Ben2021argo,2021Urban,GabrielJBrostow2008SegmentationAR,GerhardNeuhold2017TheMV,JakobGeyer2020A2D2AA} mainly focus on the \textbf{offline} performance metrics (\eg, Average Precision (AP), Truth Positive (TP)), and the model latency has not been particularly studied. Although \cite{meng2020sap,WeiHan2020StreamingOD} leverage the streaming perception paradigm \cite{meng2020sap} to measure the accuracy-latency trade-off, these benchmarks are designed for 2D detection or LiDAR-based 3D detection. 


 To address the aforementioned problem, this paper proposes the \textbf{A}utonomous-driving \textbf{S}tre\textbf{A}ming \textbf{P}erception (ASAP) benchmark. To the best of our knowledge, this is the first benchmark to evaluate the online performance of vision-centric perception in autonomous driving. The ASAP benchmark is instantiated on the camera-based 3D detection, which is the core task of vision-centric perception in autonomous driving.
To enable the streaming evaluation of 3D detectors, an annotation-extending pipeline is devised to increase the annotation frame rate of the nuScenes dataset~\cite{HolgerCaesar2019nuScenesAM} from 2Hz to 12Hz. The extended dataset, termed nuScenes-H (High-frame-rate annotation), is utilized to evaluate 3D detectors with 12Hz streaming inputs. Referring to the practical deployment, we delve into the problem of ASAP under different computational resources. Specifically, the \textbf{S}treaming \textbf{P}erception \textbf{U}nder const\textbf{R}ained-computation (SPUR) evaluation protocol is constructed: (1) To compare the model performance on varying platforms, multiple GPUs with different computation performances are assigned for the streaming evaluation. 
(2) To analyze the performance fluctuation caused by the sharing of computational resources~\cite{duato2009efficient,yeh2020kubeshare,zhang2022beverse,en2022m2bev}, the streaming evaluation is performed while the GPU is simultaneously processing other perception tasks.
As depicted in Fig.~\ref{fig:key1}, the streaming performances of different methods drop steadily as the computation power is increasingly constrained.
Besides, the model rank alters under various hardware constraints, suggesting that the offline performance cannot serve as the deterministic criterion for different approaches. Therefore, it is necessary to introduce our streaming paradigm to vision-centric driving perception.
Based on the ASAP benchmark, we further establish simple baselines for camera-based streaming 3D detection, and experiment results show that forecasting the future state of the object can compensate for the delay in inference time. Notably, the proposed BEVDepth-Sv improves the streaming performance
(mAP-S) by $\sim$2\%, $\sim$3\%, and $\sim$16\% on three GPUs (RTX3090, RTX2070S, GTX1060).

The main contributions are summarized as follows: (1) We propose the ASAP benchmark to quantitatively evaluate the accuracy-latency trade-off of camera-based perception methods, which takes a step towards the practical vision-centric perception in autonomous driving.
    (2) An annotation-extending pipeline is proposed to annotate the 12Hz raw images of the popular nuScenes dataset, which facilitates the streaming evaluation on camera-based 3D detection.
    (3) Simple baselines are established in the ASAP benchmark, which alleviates the influence of inference delay and consistently improves the streaming performances across different hardware.
    (4) The SPUR evaluation protocol is constructed to facilitate the evaluation of practical deployment, where we investigate the streaming performance of the proposed baselines and seven modern camera-based 3D detectors under various computational constraints.

\section{Related Work}
\label{sec:rw}
\subsection{Autonomous-Driving Benchmark}
Thanks to the release of various benchmarks \cite{HolgerCaesar2019nuScenesAM,PeiSun2020ScalabilityIP,Geiger2012CVPR,MariusCordts2016TheCD,bdd100k,XinyuHuang2020TheAO,Ben2021argo,GabrielJBrostow2008SegmentationAR,GerhardNeuhold2017TheMV,ShanshanZhang2017CityPersonsAD,2021Urban,JakobGeyer2020A2D2AA}, the last decade has witnessed immense progress on autonomous-driving perception. Among these benchmarks, CamVid \cite{GabrielJBrostow2008SegmentationAR}, Cityscapes \cite{MariusCordts2016TheCD}, Mapillary Vistas \cite{GerhardNeuhold2017TheMV}, Apolloscape \cite{XinyuHuang2020TheAO}, BDD100K \cite{bdd100k} and CityPerson \cite{ShanshanZhang2017CityPersonsAD} focus on the 2D annotation (segmentation masks or detection boxes). To facilitate 3D perception in autonomous driving, several benchmarks \cite{Geiger2012CVPR,AbhishekPatil2019TheHD,PeiSun2020ScalabilityIP,HolgerCaesar2019nuScenesAM,MingFangChang2019Argoverse3T,2021Urban,JakobGeyer2020A2D2AA} collect multi-modal data (RGB images, LiDAR, RADAR, GPS/IMU) with comprehensive 3D annotations.
Among these 3D annotated dataset, nuScenes \cite{HolgerCaesar2019nuScenesAM}, Waymo \cite{PeiSun2020ScalabilityIP}, Argoverse \cite{MingFangChang2019Argoverse3T}, A2D2 \cite{JakobGeyer2020A2D2AA}, Lyft L5 \cite{2021Urban} provide surround-view image data, which boosts camera-based 3D perception. Especially in the nuScenes dataset \cite{HolgerCaesar2019nuScenesAM}, a vision-centric 3D perception trend \cite{li2022bevformer,huang2021bevdet,huangjj2021BEVDet4D,liu2022petr,liu2022petrv2,li2022bevdepth,li2022bevstereo,en2022m2bev,Jin2022time,Zeng2022sts} demonstrates that camera-based 3D detectors can achieve promising accuracy. Nevertheless, these benchmarks evaluate perception methods in an offline manner, neglecting the inference time delay.

\subsection{Vision-Centric Driving Perception}
Compared with the costly LiDAR, cameras can be deployed with much lower budgets. Besides, cameras-based methods own the desirable merits to extract rich semantic information from dense color and texture information \cite{survey_bev,survey_bev_hy}, which facilitates versatile vision-centric perception in autonomous driving.
\eg, 3D detection \cite{wang2021fcos3d,wang2021pgd,MonoFlex,imvoxelnet,smoke,wang2022dfm}, semantic map construction \cite{vpn,VDE,DBLP:conf/eccv/SchulterZJC18}, and depth estimation \cite{wang2022crafting,monov2,manydepth,wei2022surround,packnet,FelixWimbauer2020MonoRecSD}. 
Recently, the Bird's Eye View (BEV) representation further promotes the development of vision-centric perception \cite{li2022hdmapnet,peng2022bevsegformer,WeixiangYang2021ProjectingYV,zou2022hft,pon,li2022bevformer,huang2021bevdet,huangjj2021BEVDet4D,liu2022petr,liu2022petrv2,li2022bevdepth,li2022bevstereo,en2022m2bev,Jin2022time,Zeng2022sts,AnthonyHu2021FIERYFI,akan2022stretchbev}.
Particularly, in the fundamental task of autonomous driving (\eg, 3D detection), BEVDet4D \cite{huangjj2021BEVDet4D}, BEVFormer \cite{li2022bevformer}, PETRv2 \cite{liu2022petrv2}, BEVDepth \cite{li2022bevdepth}, BEVStereo \cite{li2022bevstereo}, STS \cite{Zeng2022sts}, SOLOFusion \cite{Jin2022time} have achieved  promising detection accuracy, approaching that of LiDAR-based counterparts \cite{SourabhVora2019PointPaintingSF,AlexHLang2018PointPillarsFE,YinZhou2018VoxelNetEL,YanYan2018SECONDSE,yin2021center}. However, even on the powerful RTX3090, the runtime of most methods exceeds 300ms, which is far from practical deployment.

\subsection{Streaming Perception}
\label{sec:sp}
The concept of streaming perception is first proposed in \cite{meng2020sap}, where a benchmark is introduced to evaluate the accuracy-latency trade-off of 2D detectors. Faced with model latency, Kalman filter \cite{REKalman1960ANA}, dynamic scheduling \cite{meng2020sap}, and reinforcement learning \cite{AnuragGhosh2021AdaptiveSP} are utilized to alleviate the problems caused by inference time delay.
To further enhance the streaming performance, \cite{JinrongYang2022RealtimeOD} simplifies the streaming perception to the task of \textit{predicting the next frame} by an efficient detector \cite{yolox2021}.  
To investigate the streaming perception in LiDAR-based 3D detection, \cite{WeiHan2020StreamingOD} proposes to split the full-scan LiDAR points into multiple slices and process the streaming LiDAR slices with a recurrent neural network. Following the same setting, \cite{DaviFrossard2020StrObeSO} preserves past slice features and concatenates them with current slice data. \cite{QiChen2021PolarStreamSO} regards that LiDAR slices can be naturally represented in polar coordinates and the polar-pillar representation is utilized in the streaming perception. The aforementioned streaming benchmarks \cite{meng2020sap,WeiHan2020StreamingOD} are designed for 2D detection or LiDAR-based 3D detection, but the streaming paradigm of vision-centric perception in autonomous-driving (e.g., camera-based 3D detection) is still under investigation.


\begin{figure}
\centering
\resizebox{1\linewidth}{!}{
\includegraphics{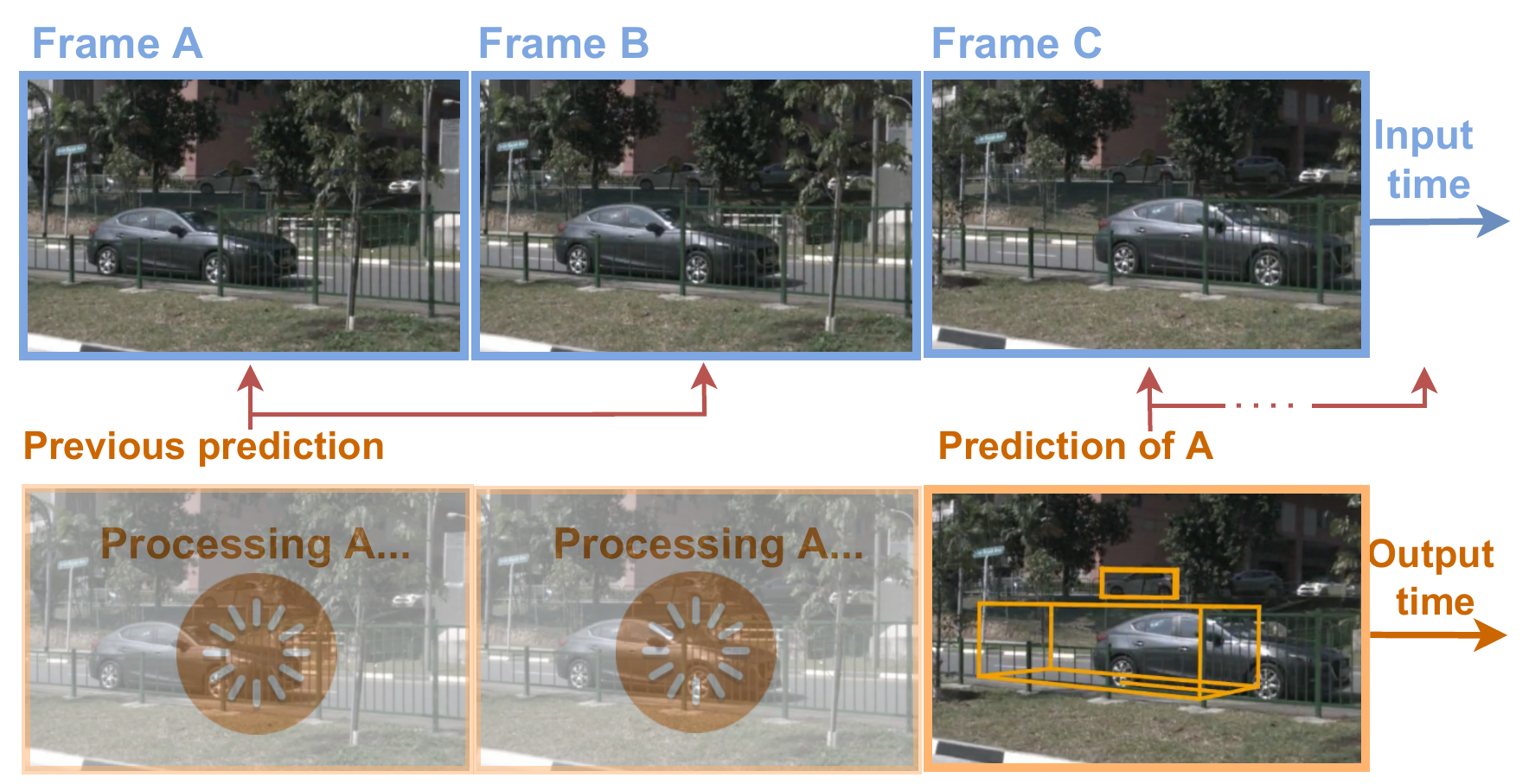}}
\caption{Illustration of the streaming evaluation in the ASAP benchmark. For \textit{every} input timestamp, the ASAP benchmark evaluates the most recent prediction if the processing of current frame is not finished.}
\label{fig:st1}
\end{figure}

\begin{figure*}[ht]
\centering
\resizebox{1\linewidth}{!}{
\includegraphics{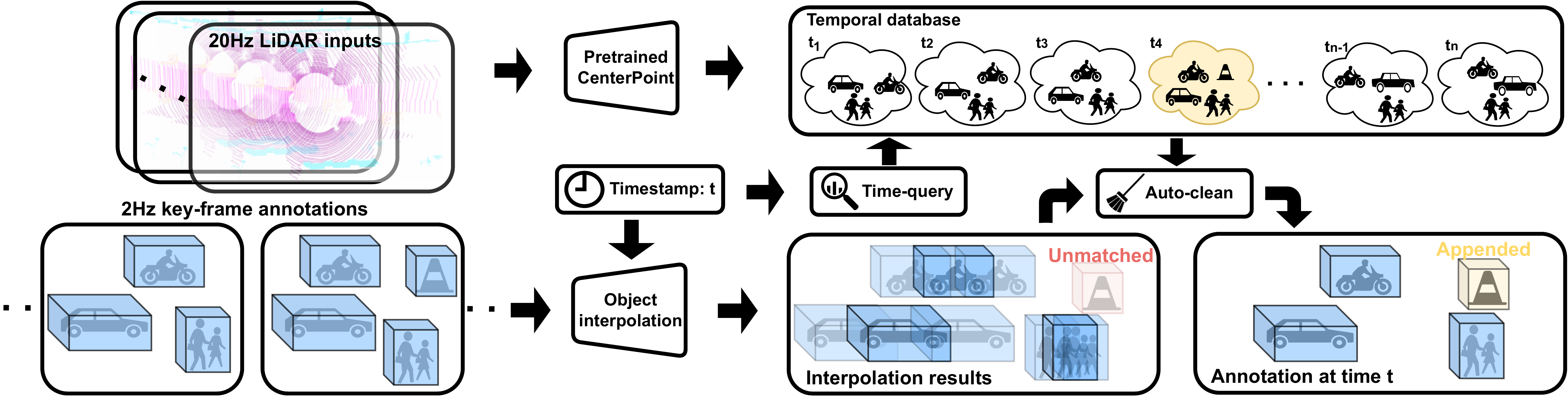}}
\caption{Overall architecture of the annotation-extending pipeline, where the 12Hz annotations are calculated by object interpolation of 2Hz key-frames, and the temporal database is established to append annotations that are missed by interpolation.}
\label{fig:ann_pipe}
\end{figure*}
\begin{figure*}[ht]
\centering
\resizebox{1\linewidth}{!}{
\includegraphics{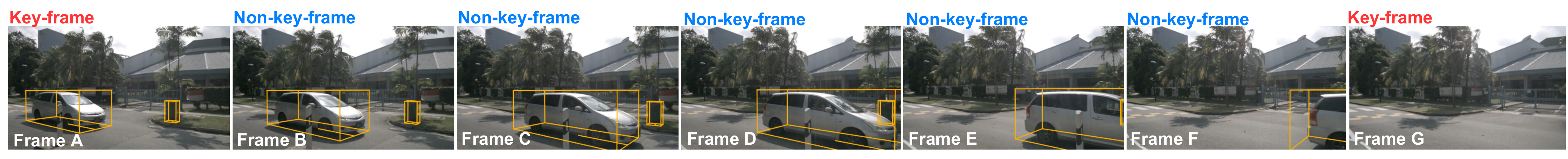}}
\caption{Visualization of the 12Hz annotated nuScenes-H dataset, where frame A and frame G are key-frames with the original 2Hz annotations, and B $\sim$ F are non-key-frames with 12Hz annotations.}

\label{fig:ann_vis}
\end{figure*}

\section{The ASAP benchmark}
\label{sec:asap}

In this section, the concept of ASAP is first introduced. Then we analyze the difficulty to evaluate streaming algorithms on the original nuScenes dataset \cite{HolgerCaesar2019nuScenesAM}, and introduce the high frame-rate nuScenes-H dataset. Subsequently, the SPUR evaluation protocol is presented to assess the streaming performance under different computational resources. Finally, we propose simple baselines to alleviate the inference time delay in streaming detection.

\subsection{Autonomous-Driving Streaming Perception}
\label{3dsap}
Referring to the streaming paradigm \cite{meng2020sap}, the ASAP benchmark conducts the evaluation in an online manner,  and the key insight is to perform the evaluation at \textit{every} input timestamp even if the processing of the current sample is not complete. Specifically, given streaming inputs $\{X_i\}_{i=1}^T$, where $X_i$ is the surround-view images at timestamp $t_i$ and $T$ is the total number of input timestamps. The perception algorithms are required to make an online response to the input instance, and the entire online predictions are $\{\hat{Y}_j\}_{j=1}^M$, where $\hat{Y}_j$ is the prediction at timestamp $t_j$, and $M$ represents the total number of predictions. Notably, the prediction timestamps are not synchronized with the input timestamps, and the model inference speed is typically slower than the input frame rate (\ie, $M<T$). To evaluate the predictions at input timestamp $t_i$, the ground truth $Y_i$ is desired to match with the most recent prediction, yielding the pair $(Y_i, \hat{Y}_{\theta(i)})$, where $\theta(i)=\arg\max\limits_{j} t_j<t_i $. Based on the matching strategy, the ASAP benchmark evaluates the online performance at \textit{every} input timestamp:
\begin{equation}
    \mathcal{L}_{\text{ASAP}} = \mathcal{L}(\{(Y_i, \hat{Y}_{\theta(i)})\}_{i=1}^{T}),
    \label{eq:metric}
\end{equation}
where $\mathcal{L}(\cdot)$ is the streaming evaluation metric, which will be elaborated in Sec.~\ref{sec:spur}. The streaming evaluation is illustrated in Fig.~\ref{fig:st1}, where the prediction of frame A is leveraged for evaluation at the timestamp of frame C. For frame A and frame B, the previous predictions are evaluated.
Notably, ASAP instantiates the streaming paradigm on camera-based 3D detection, and the key insights also generalize to other vision-centric perceptions in autonomous driving. 
\subsection{nuScenes-H}
The nuScenes dataset \cite{HolgerCaesar2019nuScenesAM} is a popular autonomous-driving perception benchmark, which significantly facilitates the vision-centric perception trend \cite{li2022bevformer,huang2021bevdet,huangjj2021BEVDet4D,liu2022petr,liu2022petrv2,li2022bevdepth,li2022bevstereo,en2022m2bev,Jin2022time,Zeng2022sts}. Consequently, it is natural to leverage the nuScenes dataset to investigate the streaming performance of various camera-based 3D detectors. However, the annotation frame rate of the original nuScenes dataset is 2Hz, which is slower than the inference speed of the majority of camera-based 3D detectors.  Therefore, it can hardly distinguish across models with different latencies. To mitigate the problem, we take advantage of the 12Hz raw images of the nuScenes dataset as the streaming input. Specifically, an annotation-extending pipeline is proposed to annotate the 12Hz raw images. The overall architecture is illustrated in Fig~\ref{fig:ann_pipe}. Given 2Hz key-frame annotations at time $t_\text{s}$ and $t_\text{e}$, we can calculate the intermediate annotation at time $t$ $(t_\text{s}<t<t_\text{e})$ using the object interpolation:
\begin{equation}
\begin{aligned}
    Tr(t) &= \frac{t_\text{e}-t}{t_\text{e}-t_\text{s}}Tr(t_\text{s}) + \frac{t-t_\text{s}}{t_\text{e}-t_\text{s}}Tr(t_\text{e}),\\
    R(t) &= \mathcal{F}_s(R(t_\text{s}), R(t_\text{e}), \frac{t_\text{e}-t}{t_\text{e}-t_\text{s}}),
\end{aligned}
\end{equation}
where $Tr(t)$ and $R(t)$ represent the object translation and rotation at time $t$, respectively. Notably, to avoid \textit{Gimbal Lock}, we employ the quaternion representation for rotation $R(t)$, and $\mathcal{F}_s$ denotes the \textit{Spherical Linear Interpolation} \cite{shoemake1985animating}.
Thanks to the instance token in the nuScenes-devkit \cite{nudev}, we can  match the corresponding objects in the sequential key-frames, which facilitates object interpolation. However, the interpolation is ignored when the object is not co-visible in sequential key-frames, thus the intermediate annotations are missed. To mitigate the problem, the temporal database $\{(t_i, Y_i^{\mathcal{L}})\}_{i=1}^n$ is established, where the predicted bounding boxes\footnote{We firstly train the CenterPoint \cite{yin2021center} on 2Hz LiDAR key-frames of the nuScenes \texttt{trainval} set, then the 3D bounding boxes of 20Hz LiDAR inputs are predicted using the CenterPoint.} $\{Y_i^{\mathcal{L}}\}_{i=1}^n$ at 20Hz input timestamps $\{t_i\}_{i=1}^n$ are stored. Then we can query the prediction $Y_{\text{query}}$ at time $t$ from the temporal database:
\begin{equation}
    Y_{\text{query}} = Y_j^{\mathcal{L}}(j=\arg\min\limits_{i}|t_i-t|).
\end{equation}
Subsequently, the auto-clean is performed by an Intersection over Union (IoU) matching between the interpolated annotation and $Y_{\text{query}}$, which filters the redundant predictions in $Y_{\text{query}}$. And the left predictions are appended to the final annotations at time $t$.
The visualization of 12Hz annotations is shown in Fig~\ref{fig:ann_vis}, where the vehicle is visible in the key-frame A, but not in the key-frame G. Hence, the intermediate annotations are ignored under the interpolation. However, such issue can be remedied by the temporal database, from which the accurate annotations are appended to the intermediate frames.
Equipped with the annotation-extending pipeline, 1M training images and 0.2M validation images are annotated in the nuScenes-H dataset. Notably, we benchmark seven 3D detectors \cite{li2022bevformer,huang2021bevdet,huangjj2021BEVDet4D,liu2022petr,li2022bevdepth, wang2021fcos3d, wang2021pgd} on the original nuScenes dataset and the extended nuScenes-H dataset, and the Pearson correlation coefficient of the two offline scores is 0.962, which validates the effectiveness of our annotation-extending method. The comparison between nuScenes-H and other autonomous-driving perception datasets \cite{Geiger2012CVPR,PeiSun2020ScalabilityIP,HolgerCaesar2019nuScenesAM,Ben2021argo,meng2020sap,WeiHan2020StreamingOD} is shown in Tab.~\ref{tab:stream_cmp}. To the best of our knowledge, nuScenes-H is the first dataset that facilitates the streaming evaluation on camera-based 3D detection.

\subsection{SPUR Evaluation Protocol}
\label{sec:spur}
Considering that the offline evaluation protocol in nuScenes \cite{HolgerCaesar2019nuScenesAM} cannot be directly adapted to the streaming system, we design the Streaming Perception Under constRained-computation (SPUR) evaluation protocol to comprehensively investigate the streaming performance of various 3D detectors.
In the next, the streaming metrics are first introduced, then the computation-constrained evaluation is elaborated.

\noindent
\textbf{Streaming metrics.} 
Average Translation Error (ATE), Average Scale Error (ASE), Average Orientation Error (AOE), Average Velocity Error (AVE), Average Attribute Error (AAE), NuScenes Detection Score (NDS) and mean Average Precision (mAP) are the official metrics in the original nuScenes dataset. These metrics can be naturally extended to the streaming metric (Eq.~\ref{eq:metric}) except for the AVE. In the streaming evaluation, the predicted bounding boxes are displaced from the ground-truth locations due to inference time delay, especially for fast-moving objects. Consequently, the majority of the anticipated true positives are slow-moving or static objects, and the AVE metric only measures the velocity error of true positive objects, which makes the streaming velocity error even lower than the offline velocity error. To address the issue, we calculate AVE as the original offline metric, and other metrics are measured with the streaming evaluation, which are termed mAP-S, ATE-S, ASE-S, AOE-S, and AAE-S. For the NDS-S, following \cite{HolgerCaesar2019nuScenesAM}, we calculate it as:
\begin{equation}
    \text{NDS-S}=\frac{1}{10}[5\text{mAP-S}+\sum\limits_{\rm{mTP}\in\mathbb{TP}}(1-\min(1,\rm{mTP}))],
\end{equation}
where $\mathbb{TP}=\{\text{AVE},\text{ATE-S},\text{ASE-S},\text{AOE-S},\text{AAE-S}\}$ is the set of true positive metrics.
\\[0.1cm]
\noindent \textbf{Computation-constrained evaluation.}
 Notably, in the ASAP benchmark, the model inference time is associated with computational resources, which influences  streaming performances. Specifically, two computation-constrained evaluation protocols are investigated: 
\begin{itemize}
    \item To compare the streaming performance on varying platforms, multiple GPUs with different performances (\eg, NVIDIA RTX3090, NVIDIA RTX 2070S, and NVIDIA GTX 1060) are assigned to evaluate 3D detectors.
    \item To analyze the performance fluctuation caused by computational resources sharing \cite{duato2009efficient,yeh2020kubeshare,zhang2022beverse,en2022m2bev}, we evaluate 3D detectors while the GPU is simultaneously processing other perception tasks (\eg, conducting $N$ classification tasks using ResNet18 \cite{he2016deep}).
\end{itemize}


\begin{table*}[t]
  \centering
  \caption{Comparison of different camera-based 3D detectors on the nuScenes-H \texttt{val} set, where the BEVDepth-Sv is the velocity-based updating baseline built upon \cite{li2022bevdepth}, and BEVDepth-Sf is the learning-based forecasting baseline built upon \cite{li2022bevdepth}. For \textit{Streaming=}\XSolidBrush, we report the 2Hz offline metrics. For \textit{Streaming=}\Checkmark, we report the streaming performance on the 12Hz ASAP benchmark.}
  	\resizebox{1\linewidth}{!}{
    \begin{tabular}{l|ccccc|ccccccc}
    \hline

    \hline
    Methods  & GPU &FPS &\#params. &GFLOPs& Streaming   & \textbf{mAP(-S)}$\uparrow$ & \textbf{NDS(-S)}$\uparrow$ & ATE(-S)$\downarrow$  & ASE(-S)$\downarrow$   & AOE(-S)$\downarrow$  & AVE$\downarrow$  &  AAE(-S)$\downarrow$   \\
    \hline
    FCOS3D   & -  &- &52.5M & 2008.2      &\XSolidBrush& 0.295  & 0.372    & 0.806             & 0.268      & 0.511       & 1.315     & 0.170            \\
    FCOS3D   & RTX3090 &  1.7 &52.5M & 2008.2  & \Checkmark & 0.208   & 0.326    & 0.828             & 0.269      & 0.512       & 1.315     &    0.175           \\
    FCOS3D   & RTX2070s &0.8 &52.5M & 2008.2 & \Checkmark & 0.151 & 0.294          & 0.836             & 0.270      & 0.522       & 1.315     & 0.187       \\
    FCOS3D   & GTX1060  &0.3 &52.5M & 2008.2  & \Checkmark & 0.051  & 0.234     & 0.858             & 0.271      & 0.585       & 1.315     & 0.200           \\
    \hline
    PGD   & -    &-   &51.3M & 2223.0    &\XSolidBrush& 0.335   & 0.409     & 0.732  & 0.263      & 0.423       & 1.285     & 0.172         \\
    PGD    & RTX3090 &  1.6 &51.3M & 2223.0  & \Checkmark & 0.206  & 0.327     & 0.817  & 0.273      & 0.488       & 1.285     & 0.185          \\
    PGD   & RTX2070s  &0.7 &51.3M & 2223.0 & \Checkmark & 0.139  & 0.289      & 0.818  & 0.276      & 0.512       & 1.285     & 0.195         \\
    PGD   & GTX1060  &0.2 &51.3M & 2223.0   & \Checkmark & 0.016    & 0.199    & 0.909  & 0.342      & 0.536       & 1.285     & 0.300        \\
    \hline
    BEVDet   & -   &-&52.6M & 215.3   &\XSolidBrush& 0.308  & 0.411     & 0.729  & 0.265      & 0.445       & 1.051     & 0.175          \\
    BEVDet   & RTX3090  &  12.6 &52.6M & 215.3 & \Checkmark & 0.289   & 0.370    & 0.730             & 0.273      & 0.533       & 1.051     &    0.209          \\
    BEVDet   & RTX2070s &8.5 &52.6M & 215.3 & \Checkmark & 0.284   & 0.367   & 0.734             & 0.273      & 0.536       & 1.051     & 0.209           \\
    BEVDet  & GTX1060&3.3 &52.6M & 215.3   & \Checkmark & 0.254    & 0.348   & 0.751             & 0.275      & 0.547       & 1.051     & 0.218           \\
    \hline
    BEVDet4D   & -   &- &53.6M & 222.0  &\XSolidBrush& 0.338  & 0.476     & 0.672  & 0.274      & 0.460       & 0.337     & 0.185         \\
    BEVDet4D  & RTX3090  &  11.9  &53.6M & 222.0 & \Checkmark & 0.309& 0.450   & 0.755  & 0.275 
         & 0.480       & 0.337     & 0.200            \\
    BEVDet4D  & RTX2070s  &  6.9 &53.6M & 222.0& \Checkmark& 0.286  & 0.438    & 0.757  & 0.275   
         & 0.481       & 0.337     & 0.201              \\
    BEVDet4D  & GTX1060   &  3.2 &53.6M & 222.0 & \Checkmark & 0.257   & 0.419  & 0.775    & 0.276
    & 0.492          & 0.337     & 0.211             \\
    \hline
    BEVFormer     & -   &-   &68.7M & 1322.2  &\XSolidBrush& 0.415 & 0.517       & 0.672  & 0.274      & 0.369       & 0.397     & 0.198       \\
    BEVFormer     & RTX3090   &  2.4  &68.7M & 1322.2  & \Checkmark &0.310   & 0.452   & 0.760  &    
    0.276      & 0.385       & 0.397     & 0.216     \\
    BEVFormer     & RTX2070s &  1.1  &68.7M & 1322.2  & \Checkmark&0.233   & 0.408  & 0.774  &    
    0.278       & 0.410          & 0.397     & 0.228           \\
    BEVFormer     & GTX1060   &  0.3 &68.7M & 1322.2 & \Checkmark & 0.074  & 0.311   & 0.819  &    
    0.280         & 0.516       & 0.397     & 0.246         \\
    \hline
    PETR               & -  & - &36.7M & 297.2     &\XSolidBrush& 0.317& 0.367      & 0.839  & 0.280 & 0.614       & 0.936     & 0.232         \\
    PETR              & RTX3090 &  6.7  &36.7M & 297.2 & \Checkmark & 0.282  & 0.341  & 0.883  & 0.288   
    &0.639         & 0.936      & 0.249           \\
    PETR              & RTX2070s  &  3.2  &36.7M & 297.2& \Checkmark& 0.254  & 0.323  & 0.897   &0.289    
    &0.658         & 0.936     & 0.258             \\
    PETR             & GTX1060  &  1.3   &36.7M & 297.2& \Checkmark & 0.195& 0.291   & 0.918   &0.291    
   & 0.659        & 0.936        & 0.266     \\
    \hline
    BEVDepth     & - &-    &76.6M & 662.6   &\XSolidBrush& 0.348  & 0.481    & 0.616  & 0.272      & 0.415       & 0.440     & 0.196         \\
    BEVDepth       & RTX3090   &  8.6  &76.6M & 662.6& \Checkmark & 0.323 & 0.464   & 0.654  & 0.272      & 0.414       & 0.440     & 0.198         \\
    BEVDepth     & RTX2070s  &  4.4 &76.6M & 662.6& \Checkmark& 0.306& 0.453   & 0.664   & 0.273     & 0.420      & 0.440     & 0.205             \\
    BEVDepth       & GTX1060   &  1.4&76.6M & 662.6 & \Checkmark & 0.226 & 0.404  & 0.686   & 0.275     & 0.449      & 0.440     & 0.235            \\
    \hline
    BEVDepth-Sv   & RTX3090   &  8.6 &76.6M & 662.6& \Checkmark & 0.328  & 0.466 & 0.654  & 0.272      & 0.416       & 0.440     & 0.198          \\
    BEVDepth-Sv   & RTX2070s & 4.4 &76.6M & 662.6 & \Checkmark& 0.316  & 0.459   & 0.662    & 0.272     & 0.419      & 0.440     & 0.198           \\
    BEVDepth-Sv   & GTX1060  &  1.4 &76.6M & 662.6 & \Checkmark & 0.263   & 0.428& 0.683   & 0.273     & 0.436      & 0.440     & 0.199            \\
    \hline
    BEVDepth-Sf   & RTX3090   &  8.6 &76.6M & 662.6& \Checkmark & 0.329  & 0.467 & 0.653  & 0.272      & 0.415       & 0.440     & 0.197          \\
    BEVDepth-Sf   & RTX2070s & 4.4 &76.6M & 662.6 & \Checkmark& 0.313  & 0.457   & 0.663    & 0.272     & 0.420      & 0.440     & 0.198           \\
    BEVDepth-Sf   & GTX1060  &  1.4 &76.6M & 662.6 & \Checkmark & 0.235   & 0.413 & 0.685   & 0.274     & 0.442      & 0.440     & 0.205            \\
    \hline
    \end{tabular}%
    }
  \label{tab:nus-val}%
\end{table*}%

\subsection{ASAP Baselines}
\begin{figure}[ht]      
\captionsetup[subfigure]{skip = 0pt}      
\centering      
\subfloat[Velocity-based updating baseline, where the Kalman filter is utilized to associate and refine multi-frame detection results, and the future state is predicted by the constant velocity motion model.]{
\resizebox{0.95\linewidth}{!}{
\includegraphics{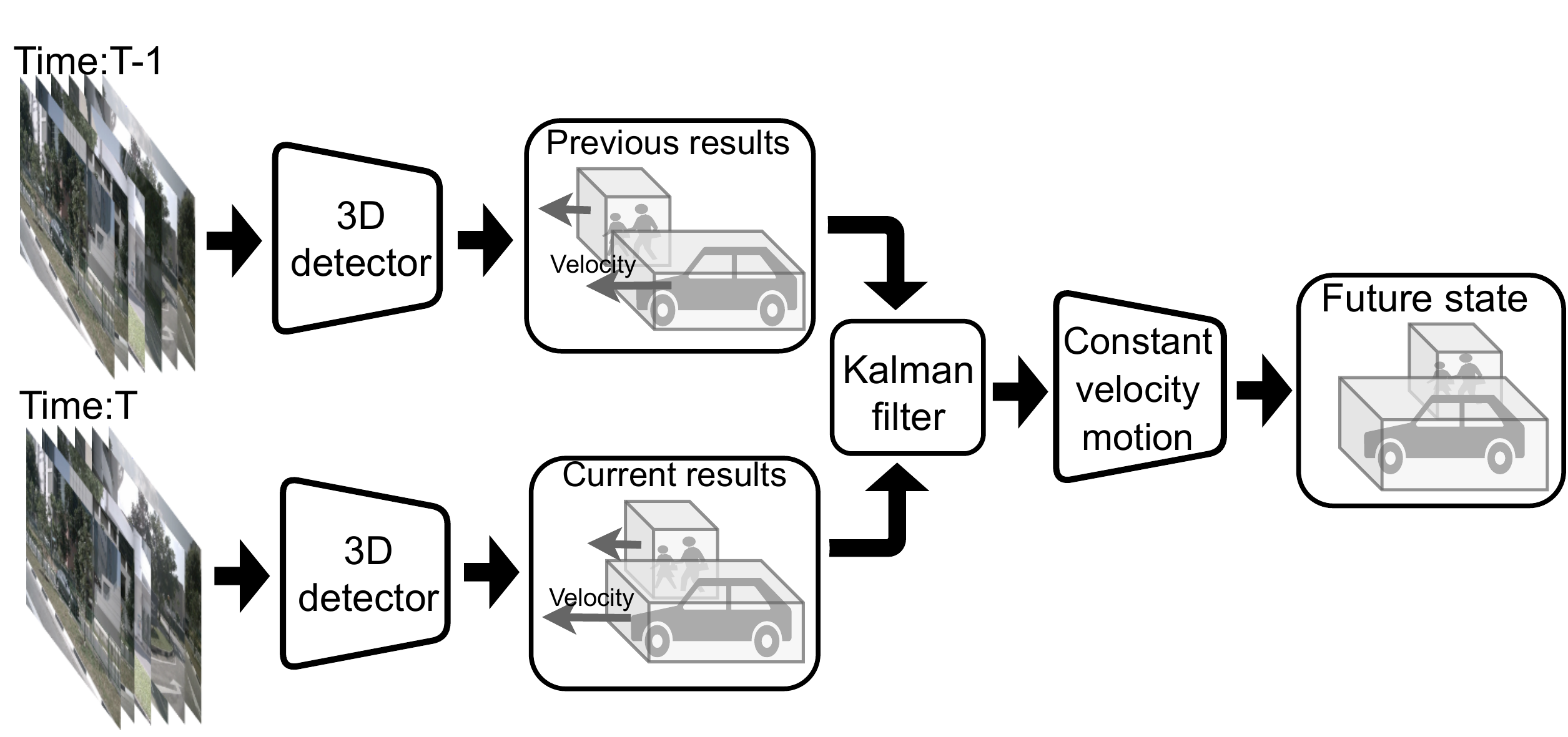}}
\label{fig:baseline1}} \\ 
\vspace{0.2cm}
\subfloat[Learning-based forecasting baseline, where the future state is directly estimated by the 3D-future detector.]
{\resizebox{0.95\linewidth}{!}{\includegraphics{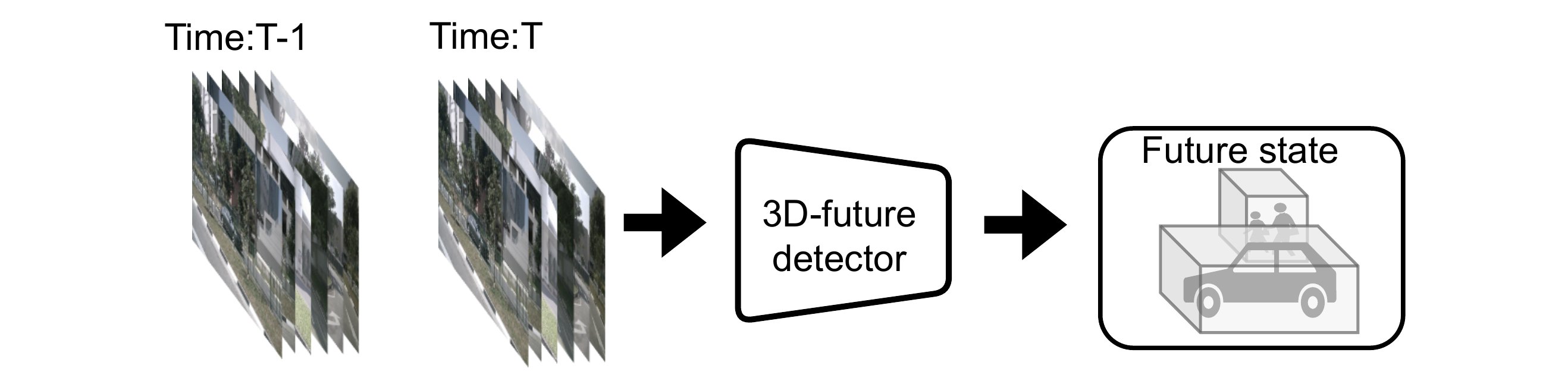}}
\label{fig:baseline2}}\ \      
\caption{Illustration of the proposed ASAP baselines.}      
\label{fig:asap_baseline}      
\end{figure}

As mentioned in Sec.~\ref{3dsap}, the ASAP benchmark evaluates the most recent predictions if the current computation is not finished, resulting in a mismatch between the previously processed observation and the current one. In this subsection, we discuss how to mitigate the mismatch problem induced by the inference time delay. Naturally, forecasting the future state emerges as a simple solution to compensate for the delay. To make future state estimations, we establish the velocity-based baseline that updates future states by the predicted object motion. Besides, we investigate a learning-based baseline that directly estimates the future locations of objects (illustrated in Fig.~\ref{fig:asap_baseline}).

\noindent
\textbf{Velocity-based updating baseline.}
Object velocity estimation is an essential task in the original nuScenes benchmark, and various 3D detectors \cite{huangjj2021BEVDet4D,li2022bevformer,li2022bevdepth,Zeng2022sts,Jin2022time,liu2022petrv2,li2022bevstereo} have been investigated to produce accurate velocity estimations. We empirically find that simply using the \textit{constant velocity motion model} can benefit streaming 3D detection:
\begin{equation}
    Tr(t_{i+1}) = Tr(t_i) + (t_{i+1}-t_i)V(t_i),
\end{equation}
where $Tr(\cdot)$ and $V(\cdot)$ represent the predicted object translation and velocity, and $t_i$, $t_{i+1}$ denote the previous input timestamp and the current evaluation timestamp. Such a velocity-based updating strategy is straightforward, but the velocity estimations are independent in each frame, neglecting that the predicted velocity should be consistent and change smoothly. To alleviate the problem, we associate predictions in consecutive frames and refine the predictions using the first-order Kalman filter \cite{REKalman1960ANA}. Specifically, IoU-based greedy matching is applied to associate 3D bounding boxes across frames. And state representation in the Kalman filter is $\{x,y,z,\dot{x},\dot{y}\}$, where $(x,y,z)$ denotes the center location of the bounding box, and $\dot{x},\dot{y}$ is the estimated velocity in the BEV plane. The updated state representations are leveraged to refine per-frame predictions. For objects that do not have correspondence in sequential frames, we simply use the \textit{constant velocity motion model} to update the locations. Notably, the velocity-based updating strategy can be applied to any modern 3D detector (\eg, in the experiment, BEVDepth-Sv is built upon BEVDepth \cite{li2022bevdepth}). Besides, the updating pipeline is lightweight ($\sim$10ms on CPU), which has a negligible impact on streaming delays.

\noindent
\textbf{Learning-based forecasting baseline.} The above baseline exploits velocity as the intermediate surrogate to predict future states. From another perspective, the streaming 3D detector should be inherently predictive of the future.
Therefore, we craft a simple framework for directly forecasting the future locations of objects. Specifically, a 3D-future detector is built upon BEVDepth \cite{li2022bevdepth} (termed BEVDepth-Sf), where the algorithm leverages history frames as input and forecasts the detection results of the \textbf{next frame}. The model architecture and training strategy are similar to those of \cite{li2022bevdepth}, except that the loss is calculated using annotations from the subsequent frame. For samples that do not have annotations for future frames, we discard them in the training phase.

\section{Streaming Evaluation on ASAP Benchmark}
In this section, the experiment setup is first given. Subsequently, we delve into the computation-constrained assessment, including streaming evaluation with platforms altering and resource sharing. Finally, we analyze the association between streaming performance and input size/backbone selection. 

\subsection{Experiment Setup}
\label{sec:exp_set}
In the ASAP benchmark, vision-centric perception is instantiated on camera-based 3D detection, which is the fundamental task in autonomous driving perception. The extended nuScenes-H dataset is leveraged to evaluate 3D detectors. Following \cite{meng2020sap}, the streaming evaluation is conducted with a hardware-dependent simulator. 
For camera-based 3D detectors \cite{wang2021fcos3d,wang2021pgd,huang2021bevdet,huangjj2021BEVDet4D,li2022bevformer,liu2022petr,li2022bevdepth} 
 evaluated in the ASAP benchmark, the model inference times are measured with their open-sourced code on a specific GPU with batch size 1. For measuring the inference time of monocular paradigms \cite{wang2021fcos3d,wang2021pgd}, we set the batch size as 6, as \cite{wang2021fcos3d,wang2021pgd} process the surround-view images independently. More implementation details are in the supplement.

\subsection{Computation-Constrained Assessment}
\noindent
Equipped with the ASAP benchmark, we analyze the streaming performance of seven modern 3D detectors (FCOS3D \cite{wang2021fcos3d}, PGD \cite{wang2021pgd}, BEVDet \cite{huang2021bevdet}, BEVDet4D \cite{huangjj2021BEVDet4D}, BEVFormer \cite{li2022bevformer}, PETR \cite{liu2022petr}, BEVDepth \cite{li2022bevdepth}) and the proposed baselines (BEVDepth-Sv, BEVDepth-Sf) under three platforms (RTX3090, RTX2070S, and GTX1060). From the results in Tab.~\ref{tab:nus-val}, it can be observed that:
\\[0.1cm]
\noindent\textbf{(1) Compared with the offline evaluation, all these 3D detectors suffer from performance drops on the ASAP benchmark.}  Even equipped with the powerful computation platform (RTX3090), the mAP-S of BEVFormer, FCOS3D and PGD relatively drop by 25.3\%, 29.5\% and 38.5\% than the offline counterparts. For efficient models (frame rate $\approx$ 10FPS) BEVDet, BEVDet4D, BEVDepth, the mAP-S still relatively drop by 6.1\%, 8.6\%, 7.2\%, as any detection results miss at least one frame in the streaming evaluation.
\\[0.1cm]
\noindent\textbf{(2) The streaming performance degrades continually as the computation power is increasingly constrained.} As the Tera Floating Point Operations Per Second (TFLOPS) alters from 35.6TFLOPS@RTX3090 to 9.1TFLOPS@RTX2070S, the  mAP-S of FCOS3D, PGD, BEVDet, BEVDet4D, BEVFormer, PETR, BEVDepth relatively drop by 27.4\%, 32.5\%, 1.7\%, 7.4\%, 24.8\%, 9.9\%, 5.3\%, respectively. Besides, when the computation performance is further constrained on 4.4TFLOPS@GTX1060, the mAP-S further decreases by a large margin. Notably, for models (FCOS3D, PGD, BEVFormer) with higher model latency, the inference frame rates are lower than 0.5FPS on GTX1060, and the corresponding mAP-S of FCOS3D, PGD, and BEVFormer are 0.051, 0.016 and 0.074, which are far from practical deployment.
\\[0.1cm]
\noindent\textbf{(3) The model rank alters under different computation performances.} An illustrative comparison is in Fig.~\ref{fig:key1}, where BEVFormer   ranks 1st in the offline evaluation and relatively outperforms the 2nd-best competitor BEVDepth by 19.3\%. However, BEVDepth suppresses BEVFormer during the streaming evaluation on RTX3090. Moreover, on the RTX2070S, the mAP-S of efficient 3D detectors (BEVDet, BEVDet4D, PETR) exceed that of BEVFormer, and the performance gap between BEVFormer and \cite{li2022bevdepth,huang2021bevdet,huangjj2021BEVDet4D,liu2022petr} is further enlarged on GTX1060.
\\[0.1cm]
\noindent
\textbf{(4) Future state estimation can compensate for the inference time delay, which improves the streaming performance.} For the velocity-based updating baseline built upon BEVDepth, BEVDepth-Sv relatively improves mAP-S by 1.5\%, 3.3\%, and 16.4\% on RTX3090, RTX2070S, and GTX1060. More baseline results are in Tab.~\ref{tab:cat}, where FCOS3D, PGD and BEVFormer relatively enhance mAP-S by 4.8\%, 5.3\% and 10.9\% on RTX3090. Note that BEVFormer-Sv obtains higher improvement due to the accurate velocity estimation (\ie, AVE@BEVFormer is 0.397, which is significantly lower than that of FCOS3D (1.315) and PGD (1.285)). Besides, we empirically find that high-speed objects particularly benefit from the velocity-based updating strategy. \eg, for BEVFormer@RTX3090, the AP-S of \textit{car} and \textit{bus} are relatively improved by 27.9\% and 37.3\%, while the relative improvement of slow-motion objects (\textit{pedestrian}) is within 5\%, which inspires future streaming algorithms to consider the discrepancy in per-category velocity. 
For the learning-based forecasting baseline built upon BEVDepth, BEVDepth-Sf relatively improves mAP-S by 1.9\%, 3.0\%, and 4.0\% on RTX3090, RTX2070S, and GTX1060. However, the improvements on RTX2070S and GTX1060 (3.0\%, 4.0\%)  are inferior to that of BEVDepth-Sv (3.3\%, 16.4\%), as the model frame rates on RTX2070S (4.4FPS) and GTX1060 (1.4FPS) are greatly slower than the streaming input speed (12FPS), and simply predicting the next frame is not sufficient to compensate for the inference time delay. This issue may be alleviated by forecasting future states of the next $N(N\ge2)$ frames, while the multi-forecasting architecture may increase the inference time and hamper the streaming perception. Therefore, end-to-end streaming 3D detection remains a promising open problem for future research. More experiment results are in the supplement.

\begin{table}[t]
  \centering
  \caption{Streaming performance (mAP-S) of FCOS3D \cite{wang2021fcos3d}, PGD \cite{wang2021pgd}, BEVFormer \cite{li2022bevformer} and the corresponding velocity-based updating baselines. The experiment are conducted on RTX3090, and we report AP-S on high-speed category (\eg, car, bus), and slow-motion category (\eg, pedestrian).}
  \resizebox{1\linewidth}{!}{
    \begin{tabular}{l|ccccc}
    \hline
    Method   & \textbf{mAP-S}$\uparrow$ &AVE$\downarrow$& Car &  Bus  & Ped.   \\

    \hline
    FCOS3D & 0.208&1.315&0.244 & 0.111   & 0.300    \\
    FCOS3D-Sv &0.218 {\color{red}(+4.8\%)}&1.315 & 0.273& 0.133   & 0.310  \\
    \hline
    PGD & 0.206&1.285&0.240 & 0.092   & 0.293     \\
    PGD-Sv &0.217 {\color{red}(+5.3\%)}&1.285& 0.266& 0.124   & 0.302  \\
    \hline
    BEVFormer & 0.310&0.397&0.373 & 0.236   & 0.402   \\
    BEVFormer-Sv &0.344 {\color{red}(+10.9\%)}&0.397& 0.477& 0.324   & 0.424  \\
    \hline
    \end{tabular}%
    }
  \label{tab:cat}%
\end{table}%

\begin{figure}[h]
\centering
\resizebox{1\linewidth}{!}{
\includegraphics{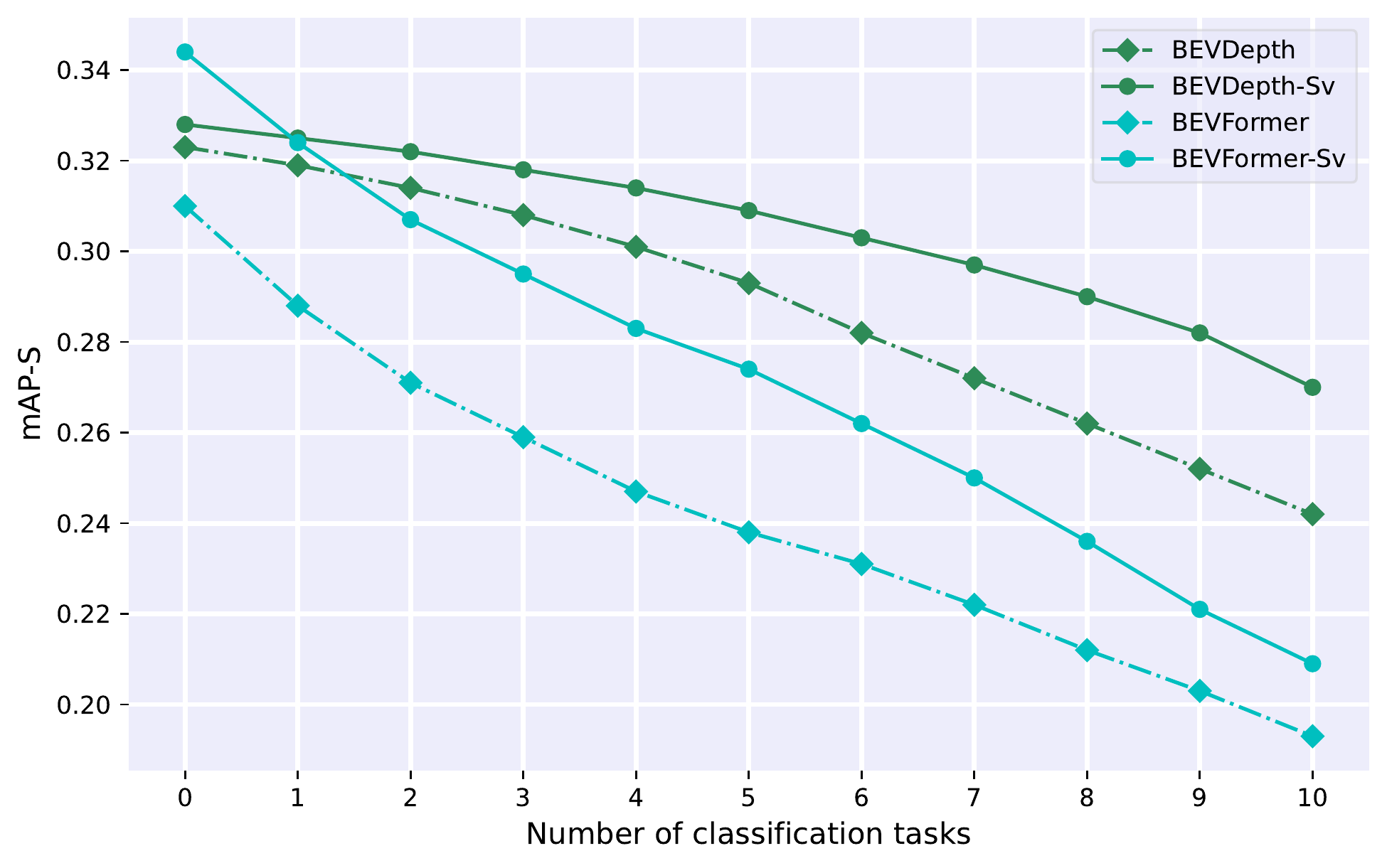}}
\caption{Comparison of streaming performance of BEVFormer  \cite{li2022bevformer}, BEVFormer-Sv, BEVDepth  \cite{li2022bevdepth} and BEVDepth-Sv under computational resources sharing, where the $x$-axis denotes the number of ResNet18-based \cite{he2016deep} classification tasks.}
\label{fig:key2}
\end{figure}

To analyze the performance fluctuation caused by computational resources sharing \cite{duato2009efficient,yeh2020kubeshare,zhang2022beverse,en2022m2bev}, we evaluate 3D detectors (BEVFormer  \cite{li2022bevformer} and BEVDepth  \cite{li2022bevdepth}) while the GPU (RTX3090) is simultaneously processing $N$ ResNet18-based \cite{he2016deep} classification tasks. As illustrated in Fig.~\ref{fig:key2}, BEVFormer and BEVDepth   suffer performance drops when the number of classification tasks increases, as fewer computational resources are allocated to the 3D detection task. Specifically, the  mAP-S of BEVFormer and BEVDepth relatively drop by 37.7\% and 25.1\% as the number of classification tasks increases from 0 to 10. Notably, the proposed velocity-based updating baselines consistently improve the streaming performance under computation sharing, and BEVDepth-Sv, BEVFormer-Sv relatively improve the mAP-S by 11.6\% and 8.3\% when 10 classification tasks are executing.

The above experiment results reveal that the computational resources significantly influence streaming performance. While the high-performance detector \cite{li2022bevformer} generates accurate predictions with powerful computation, its streaming performance suffers drops with constrained computation. In contrast, the streaming performance of efficient detectors \cite{li2022bevdepth,huang2021bevdet,huangjj2021BEVDet4D} are more consistent across different constraints, indicating that the model latency and computation budget should be regarded as design choices to optimize the practical deployment.

\subsection{Analysis on Input Size and Backbone Selection}

In this subsection, experiments are conducted to investigate the association between streaming performance and input size/backbone selection. Specifically, we evaluate BEVDepth \cite{li2022bevdepth} and BEVDepth-Sv with different image sizes ($704\times 256$, $1408\times 512$) and backbones (ResNet50, ResNet101 \cite{he2016deep}). The results are shown in Tab.~\ref{tab:modelsize}. For the offline evaluation, BEVDepth@ResNet101 relatively improves BEVDepth@ResNet50 by 3.4\%. Besides, the improvement is 18.4\% as the image resolution is further expanded ($704\times 256\rightarrow1408\times 512$). For the streaming evaluation on BEVDepth, replacing ResNet50 with ResNet101 relatively enhances mAP-S by 2.5\%. However, the mAP-S relatively drops by 3.7\% when the input size is further expanded.
Notably, our baseline BEVDepth-Sv obtains mAP-S improvement with large input size ($1408\times 512$) and ResNet101 backbone, but the improvement (4\%) is significantly lower than that of the offline metric (18.4\%). The above results inspire that high-resolution inputs and stronger backbones may hinder streaming performance due to high latency.
Therefore, the input size/backbone selection should be meticulously designed in the practical deployment.

\begin{table}[t]
  \centering
  \caption{Evaluation with different input sizes and backbones. For \textit{Streaming}=\Checkmark, we conduct the streaming evaluation on RTX3090. For \textit{Streaming}=\XSolidBrush, the offline evaluation is performed.}
  \resizebox{1\linewidth}{!}{
    \begin{tabular}{lcc|c|c}
    \hline

    \hline
    Method   & Backbone &  Input Size &Streaming& mAP(-S) $\uparrow$  \\
    \hline
    BEVDepth   & R-50 & $704\times 256$  & \XSolidBrush&0.348         \\
    BEVDepth& R-101 & $704\times 256$   & \XSolidBrush  &0.360 {\color{red}(+3.4\%)}   \\
    BEVDepth  & R-101 & $1408\times 512$   & \XSolidBrush  &0.412  {\color{red}(+18.4\%)}  \\
    \hline
    BEVDepth   & R-50 & $704\times 256$ & \Checkmark &0.323         \\
    BEVDepth  & R-101 & $704\times 256$  & \Checkmark &0.331 {\color{red}(+2.5\%)}    \\
    BEVDepth  & R-101 & $1408\times 512$ & \Checkmark  &0.311 {\color{green}(-3.7\%)}    \\
    \hline
    BEVDepth-Sv   & R-50 &  $704\times 256$ & \Checkmark &0.328         \\
    BEVDepth-Sv  & R-101 &  $704\times 256$  & \Checkmark &0.340 {\color{red}(+3.7\%)}    \\
    BEVDepth-Sv  & R-101 & $1408\times 512$  & \Checkmark &0.341 {\color{red}(+4.0\%)}    \\
    \hline
    \end{tabular}%
    }
  \label{tab:modelsize}%
\end{table}%

\section{Conclusion}
In this paper, the ASAP benchmark is proposed to evaluate the online performance of vision-centric driving perception approaches. Specifically, we extend 12Hz raw images of nuScenes dataset, and introduce the nuScenes-H dataset for camera-based streaming 3D detection. Besides, the SPUR protocol is established for computation-constrained evaluation. Additionally, we propose ASAP baselines to compensate for the inference time delay, which consistently enhance the streaming performance on three hardware. Equipped with the ASAP benchmark, we investigate the streaming performance of seven modern camera-based 3D detectors and two proposed baselines under various computation constraints. The experiment results reveal that the model latency and computation budget should be regarded as design choices for practical deployment.

\section{Limitation and Future Work} 
The proposed ASAP benchmark takes a step to practical vision-centric perception in autonomous driving. Currently, the ASAP benchmark utilizes modern GPUs (\eg,  NVIDIA RTX3090, NVIDIA RTX2070S, NVIDIA GTX1060) to conduct the streaming evaluation. A more practical strategy is to evaluate with the computation of system-on-a-chip (\eg, NVIDIA Thor, Mobileye EyeQ5, Horizon $\text{Journey} 5$), and deploy algorithms with 8-bit int/floating point (INT8, FP8) precision acceleration. Furthermore, in future work, more autonomous-driving tasks (\eg, semantic map construction, depth estimation, motion forecasting) should be considered in the SPUR evaluation protocol, towards multi-task and end-to-end autonomy.
{\small
\bibliographystyle{ieee_fullname}
\bibliography{PaperForReview}
}

\newpage
\section{Additional Implementation Details}
\subsection{Streaming Simulation}
\begin{figure}[h]
\centering
\resizebox{1\linewidth}{!}{
\includegraphics{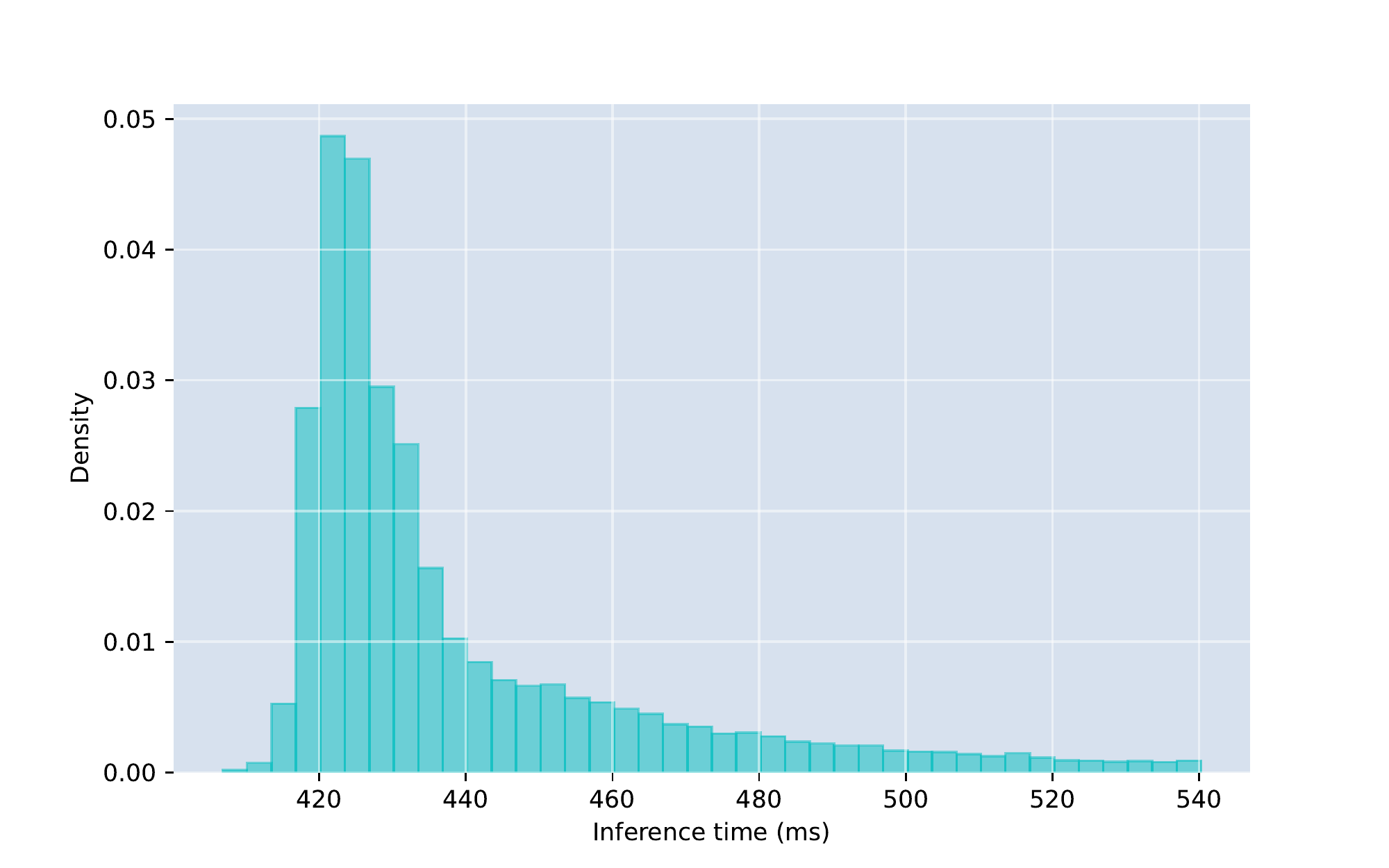}}
\caption{Inference time distribution for BEVFormer \cite{li2022bevformer} (Backbone: ResNet101 \cite{he2016deep}, Input size: $1600\times 900$)  on NVIDIA RTX3090.}
\label{fig:hist}
\end{figure}

As described in Sec.~3.1 (main text), to evaluate the predictions $\hat{Y}$ at input timestamp $t_i$, the ground truth $Y_i$ is desired to match with the most recent prediction, yielding the pair $(Y_i, \hat{Y}_{\theta(i)})$, where $\theta(i)=\arg\max\limits_{j} t_j<t_i $. The input time $\{t_i\}_{i=1}^T$ is a 12Hz sequence, but the output time $\{t_j\}_{j=1}^M$ of each prediction is associated with the model runtime on specific hardware. 
To determine the output timestamps, the streaming evaluation is conducted with a hardware-dependent simulator \cite{meng2020sap}. Specifically, we run the algorithm over the entire nuScenes \cite{HolgerCaesar2019nuScenesAM}, and measure the inference time of the algorithm on a specific GPU (the runtime distribution of BEVFormer \cite{li2022bevformer} on NVIDIA RTX3090 is shown in Fig.~\ref{fig:hist}). Then we can randomly sample model runtime from the time distribution, to calculate the output timestamps $\{t_j\}_{j=1}^M$ in the simulation.


\subsection{Streaming Evaluation Details}
In the ASAP benchmark, we analyze the streaming performance of seven modern 3D detectors. We use their open-sourced code and pretrained model (BEVDet-Tiny \cite{bevdet_git}, BEVDet4D-Tiny \cite{bevdet4D_git}, BEVFormer-Base \cite{bevformer_git}, BEVDepth-R50 \cite{bevdepth_git}, PETR-R50 \cite{petr_git}, FCOS3D-R101 \cite{fcos3d_git}, PGD-R101 \cite{pgd_git}) to generate detection results from the 12Hz streaming inputs. Notably, for multi-frame methods (\eg, BEVFormer \cite{li2022bevformer}, BEVDepth \cite{li2022bevdepth}) that use sequential frames as input, we set the input-frame-interval as six (instead of one in the original 2Hz input configuration). Such a strategy maintains the input-timestamp-interval as 0.5s, which guarantees sufficient \textit{Triangulation Priority} \cite{JohannesLSchonberger2016PixelwiseVS} for 3D perception. As shown in Tab.~\ref{tab:offline12}, for BEVFormer and BEVDepth, the proposed configuration (input-frame-rate (I.F.I)=6, input frequency (I.F.)=12Hz) significantly outperforms the original setting (I.F.I=1, I.F.=12), and the corresponding metrics (mAP, ATE, ASE, AOE, AVE, AAE) are comparable to those of the 2Hz result (I.F.I=1, I.F.=2).

\begin{table}[ht]
  \centering
  \caption{Offline performance of BEVFormer \cite{li2022bevformer} and BEVDepth \cite{li2022bevdepth} on the nuScenes (I.F.=2Hz) and nuScenes-H (I.F.=12Hz) , where I.F. represents the input frequency, and I.F.I denotes the input-frame-interval.}
  \resizebox{1.0\linewidth}{!}{
    \begin{tabular}{l|cc|cccccc}
    \hline
    Method & I.F (Hz)& I.F.I & \textbf{mAP}$\uparrow$ &ATE$\downarrow$&ASE$\downarrow$&AOE$\downarrow$ &AVE$\downarrow$ &AAE$\downarrow$\\ 
    \hline
    BEVFormer & 2 &  1  &  0.415 & 0.672 & 0.274 & 0.369 & 0.397 & 0.198 \\
    BEVFormer & 12 &  1  &  0.341 & 0.769 & 0.279 & 0.400 & 0.699 & 0.203 \\
    BEVFormer & 12 &  6  &  0.410 & 0.691 & 0.274 & 0.376 & 0.401 & 0.197 \\
    \hline
    BEVDepth & 2 &  1  &  0.348 & 0.616 & 0.272 & 0.415 & 0.440 & 0.196 \\
    BEVDepth & 12 &  1  &  0.311 & 0.640 & 0.274 & 0.470 & 0.893 & 0.209 \\
    BEVDepth & 12 &  6  &  0.341 & 0.622 & 0.273 & 0.412 & 0.453 & 0.193 \\
    \hline

    \end{tabular}%
    }
  \label{tab:offline12}%
\end{table}%

\begin{table}[ht]
  \centering
  \caption{Streaming performance (mAP-S) of FCOS3D \cite{wang2021fcos3d}, PGD \cite{wang2021pgd}, BEVFormer \cite{li2022bevformer}, BEVDet \cite{huang2021bevdet}, BEVDet4D \cite{huangjj2021BEVDet4D}, PETR \cite{liu2022petr} and the corresponding velocity-based updating baselines. The experiments are conducted on RTX3090.}
  \resizebox{1\linewidth}{!}{
    \begin{tabular}{l|ccccc}
    \hline
    Method   & \textbf{mAP-S}$\uparrow$ &ATE-S$\downarrow$&ASE-S$\downarrow$&AOE-S$\downarrow$ &AAE-S$\downarrow$   \\

    \hline
    FCOS3D & 0.208&0.828&0.268&0.511&0.170   \\
    FCOS3D-Sv &0.218 {\color{red}(+4.8\%)}& 0.820&0.267&0.506&0.169 \\
    \hline
    PGD & 0.206&0.817&0.273&0.488&0.185     \\
    PGD-Sv &0.217 {\color{red}(+5.3\%)}& 0.813&0.273&0.485&0.183\\
    \hline
    BEVFormer & 0.310&0.760&0.276&0.385&0.216  \\
    BEVFormer-Sv &0.344 {\color{red}(+10.9\%)}& 0.748 &0.274& 0.382& 0.208  \\
    \hline
    BEVDet & 0.289&0.730&0.273&0.533&0.209 \\
    BEVDet-Sv &0.291 {\color{red}(+0.7\%)}& 0.728 & 0.273 &0.532&0.207 \\
    \hline
    BEVDet4D & 0.309&0.755&0.275&0.480&0.200 \\
    BEVDet4D-Sv &0.316 {\color{red}(+2.3\%)}& 0.750 & 0.274 &0.476&0.198 \\
    \hline
    PETR & 0.282&0.883&0.288&0.639&0.249 \\
    PETR-Sv &0.291 {\color{red}(+3.2\%)}& 0.880 & 0.287 &0.636&0.247 \\
    \hline
    \end{tabular}%
    }
  \label{tab:velmore}%
\end{table}%

\begin{table}[h]
  \centering
  \caption{Ablation study of the velocity-based updating baseline, where \textit{C.V.} represents the \textit{constant velocity motion model}, and $K.F.$ denotes the Kalman filter refinement. The streaming evaluation is conducted on RTX3090.}
  \resizebox{1\linewidth}{!}{
    \begin{tabular}{l|cc|cccc}
    \hline

    \hline
    Methods & C.V.  & K.F.   & mAP-S $\uparrow$ & NDS-S$\uparrow$ &ATE-S $\downarrow$&AOE-S $\downarrow$\\
    \hline
    BEVFormer  & &           &0.310    & 0.452 & 0.760 & 0.385      \\
    BEVFormer  &\checkmark  &    &0.332  & 0.460 & 0.756 & 0.384   \\
    BEVFormer  &\checkmark  & \checkmark   &0.344  & 0.465 & 0.748 & 0.382   \\
    \hline
    
    FCOS3D  & &           &0.208    & 0.326 & 0.828 & 0.512     \\
    FCOS3D  &\checkmark  &    &0.212  & 0.329 & 0.823 & 0.509   \\
    FCOS3D  &\checkmark  & \checkmark   &0.218  & 0.332 & 0.820 & 0.506   \\
    \hline

    \end{tabular}%
    }
  \label{tab:basevel}%
\end{table}%

\section{Additional Baseline Results}
In this section, we provide additional experiment results of the velocity-based updating baseline.
As shown in Tab.~\ref{tab:velmore}, the proposed baselines built upon \cite{wang2021fcos3d,wang2021pgd,li2022bevformer,huang2021bevdet,huangjj2021BEVDet4D,liu2022petr} consistently enhance the streaming performance, suggesting that the velocity-based updating baseline can compensate for the inference delay. Note that BEVDet-Sv and BEVDet4D-Sv obtain relatively lower improvements than other methods, as they suffer little from the influence of inference delay. Namely, the model speed of BEVDet@RTX3090 and BEVDet4D@RTX3090 are $\sim$12Hz, which is close to the input frame rate.

Besides, we conduct ablation study to validate the effectiveness of the Kalman filter refinement. As shown in Tab.~\ref{tab:basevel}, FCOS3D \cite{wang2021fcos3d} and BEVFormer \cite{li2022bevformer} relatively improve the mAP-S by 4.8\% and 7.1\% using the \textit{constant velocity motion model}. Notably, the Kalman filter further boosts the mAP-S (11.0\%@BEVFormer and 9.1\%@FCOS3D), which indicates that multi-frame association and state refinement can benefit the streaming perception.

\section{Visualizations}
\begin{figure}[t]
\centering
\resizebox{0.9\linewidth}{!}{
\includegraphics{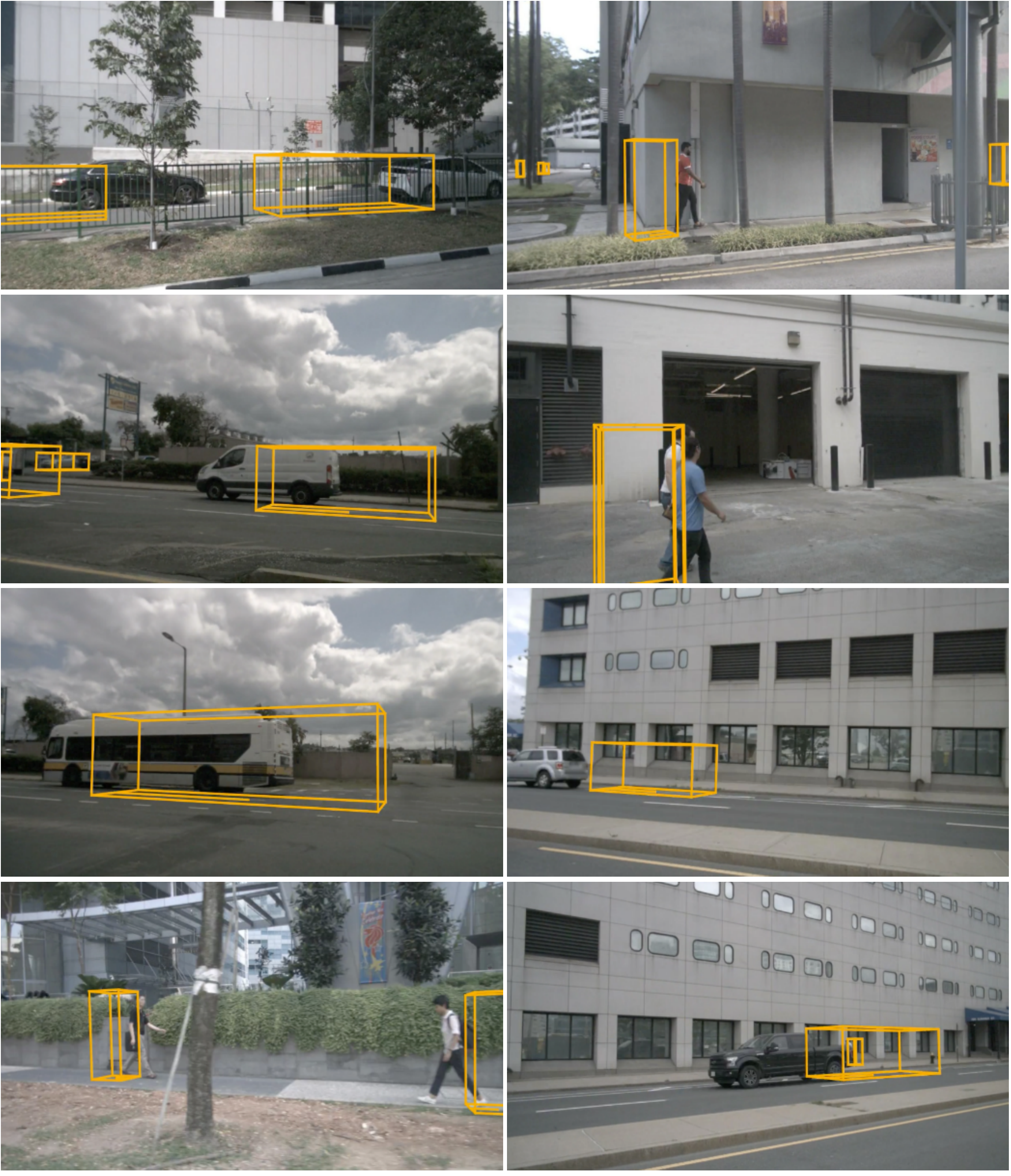}}
\caption{Visualization of the streaming perception results, where the predicted bounding boxes are displaced from the moving objects (\eg, car, pedestrian).}
\label{fig:st_6}
\end{figure}

\begin{figure}[h]
\centering
\resizebox{0.9\linewidth}{!}{
\includegraphics{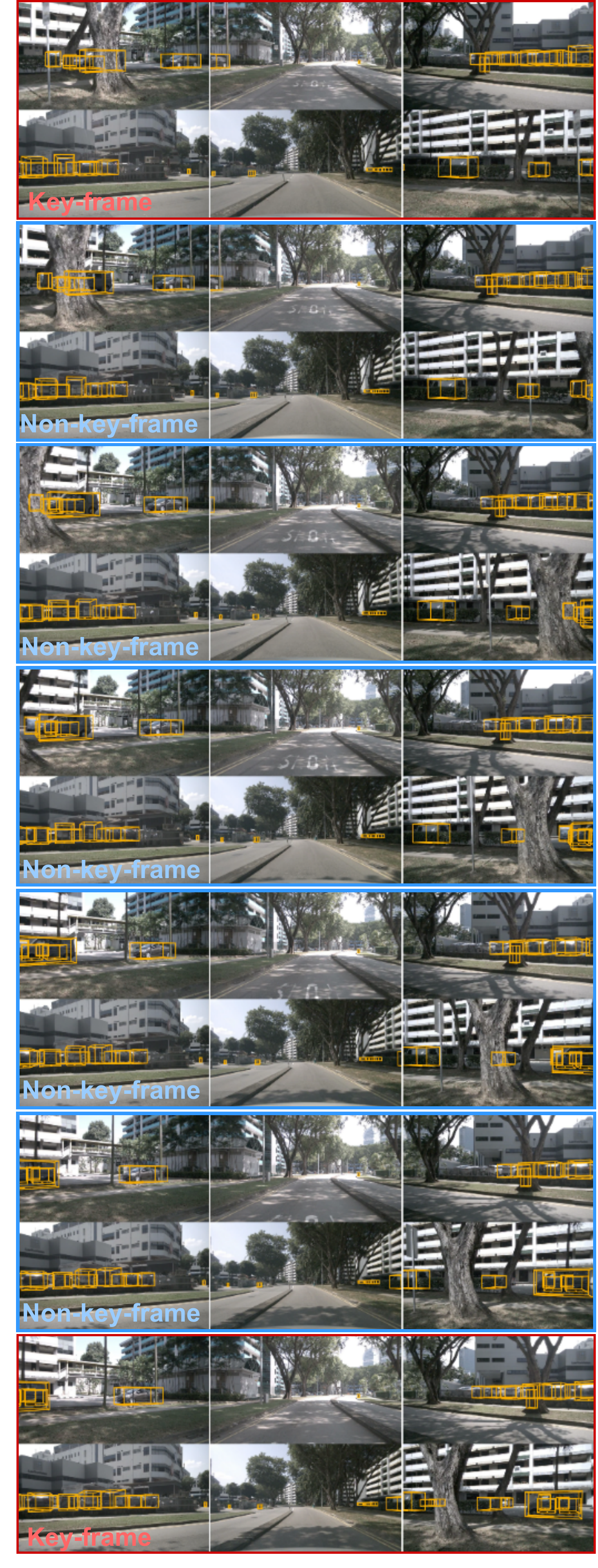}}
\caption{Visualization of the surround-view annotation in nuScenes-H, where the key-frames are the 2Hz images in the original nuScenes dataset \cite{HolgerCaesar2019nuScenesAM}, and the intermediate non-key-frames are the annotated 12Hz images.}
\label{fig:vis-12Hz}
\end{figure}

As depicted in Fig.~\ref{fig:vis-12Hz}, we visualize the 12Hz annotations of nuScenes-H (more visualization comparison between the 2Hz nuScenes and 12Hz nuScenes-H can be found in the uploaded video files). Besides, the streaming detection results are shown in Fig.~\ref{fig:st_6}, where the predicted bounding boxes are displaced from the object locations, especially for the high-speed vehicles.

\end{document}